%% file: journal.tex
\documentclass[journal]{IEEEtran}
\IEEEoverridecommandlockouts
\usepackage{cite}
\usepackage{amsmath,amssymb,amsfonts}
\usepackage{algorithmicx}
\usepackage{graphicx}
\usepackage{textcomp}
\usepackage{xcolor}
\usepackage{comment}
\def\BibTeX{{\rm B\kern-.05em{\sc i\kern-.025em b}\kern-.08em
    T\kern-.1667em\lower.7ex\hbox{E}\kern-.125emX}}

\usepackage[shortcuts,acronym]{glossaries}
\newacronym{rl}{RL}{Reinforcement Learning}
\newacronym{dl}{DL}{Deep Learning}
\newacronym{drl}{DeepRL}{Deep Reinforcement Learning}
\newacronym{irl}{IRL}{Interactive Reinforcement Learning}
\newacronym{dirl}{DeepIRL}{Deep Interactive Reinforcement Learning}
\newacronym{mdp}{MDP}{Markov Decision Processes}
\newacronym{cnn}{CNN}{Convolutional Neural Network}
\newacronym{ppr}{PPR}{Probabilistic Policy Reuse}
\newacronym{bpa}{BPA}{Broad-persistent Advising}

\graphicspath{{figures/}}
\usepackage{array}
\usepackage{booktabs} 
\usepackage{graphicx}
\usepackage{subcaption}
\usepackage{algorithm}
\usepackage{algpseudocode}

\begin{document}

\title{A Broad-persistent Advising Approach for Deep Interactive Reinforcement Learning in Robotic Environments}


\author{Hung~Son~Nguyen,
        Francisco~Cruz,
        and~Richard~Dazeley

\thanks{The authors are with the School of Information Technology, Deakin University, Geelong, Australia (e-mails:
hsngu@deakin.edu.au; francisco.cruz@deakin.edu.au;
richard.dazeley@deakin.edu.au)}

}

\maketitle

\begin{abstract}
\ac{drl} methods have been widely used in robotics to learn about the environment and acquire behaviors autonomously. \ac{dirl} includes interactive feedback from an external trainer or expert giving advice to help learners choosing actions to speed up the learning process. However, current research has been limited to interactions that offer actionable advice to only the current state of the agent. Additionally, the information is discarded by the agent after a single use that causes a duplicate process at the same state for a revisit. In this paper, we present \ac{bpa}, a broad-persistent advising approach that retains and reuses the processed information. It not only helps trainers to give more general advice relevant to similar states instead of only the current state but also allows the agent to speed up the learning process. We test the proposed approach in two continuous robotic scenarios, namely, a cart pole balancing task and a simulated robot navigation task. The obtained results show that the performance of the agent using \ac{bpa} improves while keeping the number of interactions required for the trainer in comparison to the \ac{dirl} approach.
\end{abstract}


\input{components/01_introduction}
\input{components/02_related_work}

\input{components/03_approach}
\input{components/04_experimental}
\input{components/05_results}
\input{components/06_conclusion}

\bibliographystyle{IEEEtran}
\bibliography{references}

\begin{IEEEbiography}[{\includegraphics[width=1in,height=1.25in,clip,keepaspectratio]{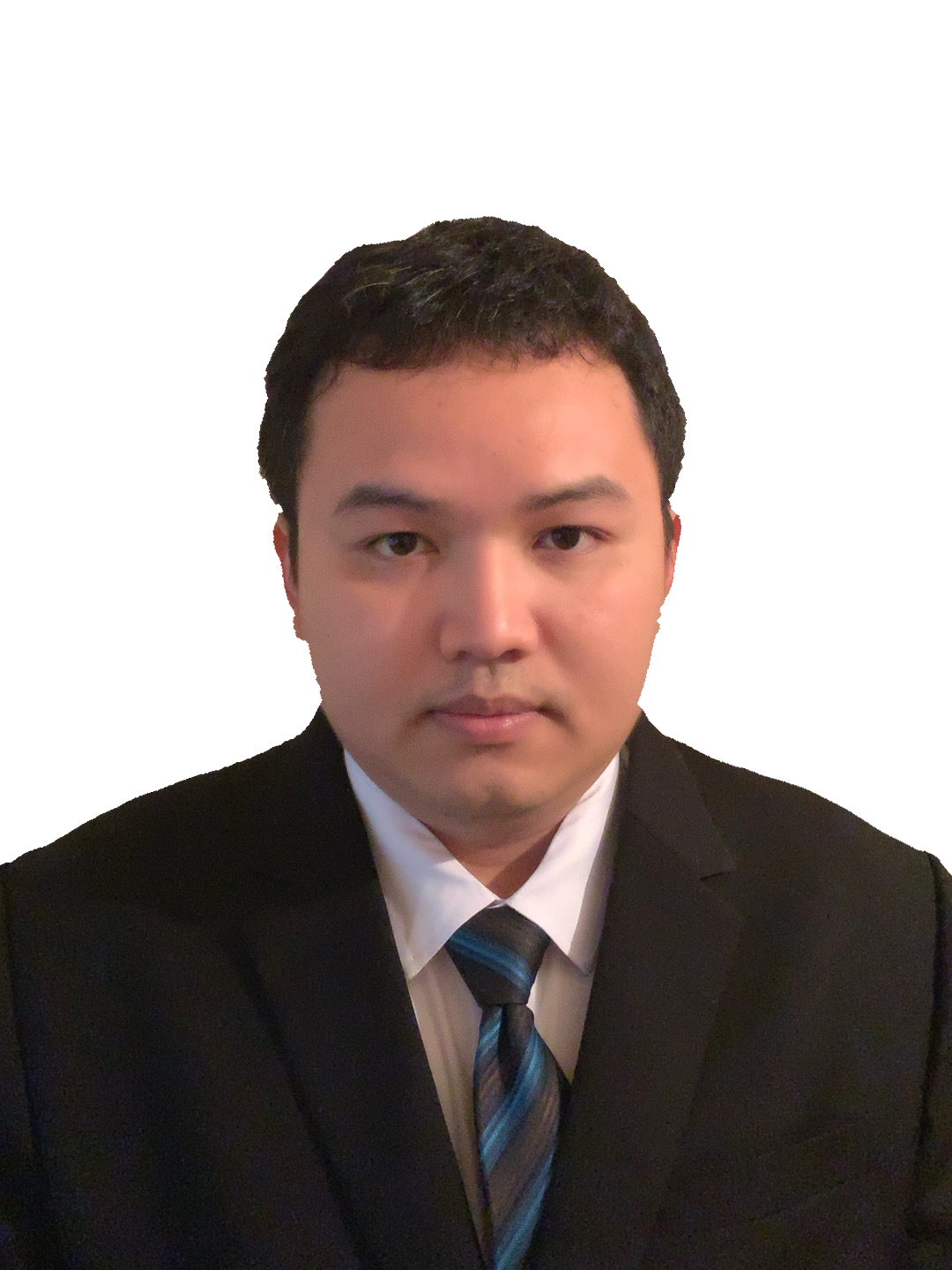}}]{Hung Son Nguyen}
received the Bachelor of Science degree in field of information technology and the Master of Science degree in computer science from the University of Science - Vietnam National University, Ho Chi Minh, Vietnam in 2010  and  2016, respectively. 

He is currently pursuing a master's degree in artificial intelligence at School  of  IT, Deakin University, Geelong, Victoria, Australia from 2020. His current research interests include computer vision, artificial neural networks, deep learning, reinforcement learning and human-robot interaction.

\end{IEEEbiography}

\vskip 0pt plus -1fil

\begin{IEEEbiography}[{\includegraphics[width=1in,height=1.25in,clip,keepaspectratio]{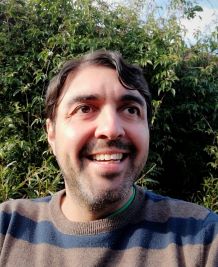}}]{Francisco Cruz}
received the bachelor’s degree in engineering and the master’s degree in computer engineering from the University of Santiago, Chile, in 2004 and 2006, respectively, and the Ph.D. degree from the University of Hamburg, Germany, in 2017, working in developmental robotics focused on interactive reinforcement
learning.

In 2015, he was a Visiting Researcher with the Emergent Robotics Laboratory, Osaka University. He joined the Engineering School, Universidad Central de Chile, as a Research and Teaching Associate, in 2017, and the School of IT, Deakin University, as a Research Fellow, in 2019. His current research interests include reinforcement learning, explainable artificial intelligence, human–robot interaction, artificial neural networks, and psychologically and bio-inspired models.
\end{IEEEbiography}

\vskip 0pt plus -1fil

\begin{IEEEbiography}[{\includegraphics[width=1in,height=1.25in,clip,keepaspectratio]{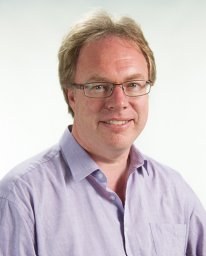}}]{Richard Dazaley}
received the Ph.D. degree in computer science from the University of Tasmania, in 2007. He is currently an Associate Professor of computer science with Deakin University, Geelong, where he is also the Deputy Leader of the Machine Intelligence Lab. He is widely recognized
for his pioneering work in multiobjective reinforcement learning publishing several highly cited papers that helped established the field. He has also organized multiple workshops in the field.
More recently, he has been interested in applying reinforcement learning and multiobjective principles in interactive, safe, explainable, and secure systems. He is also interested in prudence analysis, natural language processing, data analytics, and security. He has received multiple awards including the Australian Educator of the Year from the Australian Computer Society,
in 2016
\end{IEEEbiography}
\end{document}

%% file: components/01_introduction.tex
\section{Introduction}
Robot development has achieved big steps of improvement and gains more attention in recent years. This success does not only come from industrial areas where robots are gradually replacing humans \cite{Dahlin2019}, but also known in the domestic areas. Their presence in domestic environments is still limited, mainly due to the presence of many dynamic variables \cite{Cruz2019} and safety requirements \cite{Tadele2014}. Intelligence robots in the future should be able to know and detect users, learn action objects, select opportunities, and learn to behave in domestic scenarios. To successfully perform these complex tasks, robots face many challenges such as pattern recognition, navigation and object manipulation all in different environmental conditions. That is, robots in the domestic environment need to be able to continuously acquire and learn new skills.

\acrfull{rl} is a method used for a robot controller in order to learn optimal policy through interaction with the environment, through trial and error \cite{Sutton2018}. The use of \acrshort{rl} in previous results shows that there is great potential for using \acrshort{rl} in robots \cite{Cruz2019}. Especially, \acrfull{drl} has also achieved promising results in manipulation skills \cite{Wang2020, Nguyen2019}, and on how to grasp as well as legged locomotion \cite{Ibarz2021}. However, there is an open issue relating to the performance in the \acrshort{rl} and \acrshort{drl} algorithms, which is the excessive time and resources required by the agent to achieve acceptable outcomes \cite{millan2019human, Ayala2019}. The larger and complex the state space is, the more computational costs will be spent to find the optimal policy. 

Among of different approaches to speed up this process, there is one promising method named \acrfull{irl} that can improve convergence speed and has showed its feasibility. \acrshort{irl} allows a trainer to give advice or evaluate a learning agent’s behaviour  \cite{Cruz2016}, helps the agent to shape exploration policy and reducing search space in the early stages. Combining  \acrshort{irl} with \acrshort{drl} gives a model of \acrfull{dirl} which can be used in continuous space with improved learning speed \cite{Moreira2020}. However, current techniques using \acrshort{dirl} allow trainers to evaluate or recommend actions based only on the current state of the environment. The advice from the trainer have discarded causes by the agent after a single use, which leads to duplicate process at the same state for reuse.

This work introduces the \acrfull{bpa} approach for \acrshort{dirl} to provide the agent a method for information retention and reuse of previous advice from a trainer. This approach includes two components: generalisation and persistence.  In this article, we used $k$-means algorithm and \acrfull{ppr} for each component, respectively. Agents using the \acrshort{bpa} approach have better results than their non-using counterparts while keeping the number of interactions required for the trainer.

%% file: components/02_related_work.tex
\section{Related works}
\subsection{Deep reinforcement learning}
\acrfull{rl} is a branch of machine learning in which artificially intelligent agents learn behaviours by interacting with their surroundings \cite{Sutton2018}. Reinforcement learning tools learn through trial and error by repeatedly interacting with the surrounding environment and learning which actions do and which actions will not produce the expected results. 

\acrshort{rl} is appropriate for studying tasks that may be modelled as \acrfull{mdp} \cite{Sutton2018}. An \acrshort{mdp} is specified by the tuple ($S$,$A$,$T$,$R$, $\gamma$), where $S$ is a finite set of states in the environment, $A$ is a set of actions available in each state, $T$ is the transition function $T$ : $S_n \times A \rightarrow S_{n+1}$, $R$ is the reward function $R$ : $S \times A \rightarrow R$, and $\gamma$ is a discount factor which is $0 \leq \gamma \leq 1$

In the \acrshort{rl} setup, a machine learning algorithm-controlled agent observes a state $s_t$ from its environment at timestep $t$. In state $s_t$, the agent communicates with the environment by performing action $a_t$. Then the agent moves to a new state $s_{t+1}$ and receive reward $r_{t+1}$ as feedback from environment based on the previous state and the chosen action. Therefore, the reward collected by policy $\pi$ at timestep $t$ is shown in Equation (\ref{eq:1})

\begin{equation} \label{eq:1}
    r_t + \gamma r_{t+1} + \gamma^2 r_{t+2} + ... = \sum_{k = 0}\gamma^k r_{t+k}
\end{equation}

where $r_t$ is the reward at timestep $t$. The discount rate $\gamma$ stands for the importance of rewards in the future. The agent's aim is to find out a policy $\pi$ that maximises anticipated profit (reward).

In conventional \acrshort{rl} algorithms, most of the time, are only considered \acrshort{mdp} with discrete states and actions space. However, in many real-world applications, the state space is not really discrete, but rather a continuous domain \cite{dulac2019challenges, millan2021robust}. Therefore, to be usable in the continuous state space, neural networks are also considered as function approximators that are especially useful in \acrshort{rl} when the state space or action space is too broad to fully comprehend \cite{mnih2013playing}. Neural nets can discover ways to map states to values in this way. When the problem state space is too big or considered as continuous space, we cannot use a lookup table to store and update all possible states and actions. In that case, one alternative is to train a neural network with samples from the state and the environment and expect them to predict the value of the next action as our target in \acrshort{rl}. More formally, we use a neural network to approximate the optimal action-value function which is the maximum sum of rewards in Equation (\ref{eq:2})

\begin{equation} \label{eq:2}
    Q^{*}(s_t,a_t) =  \max_{\pi}\mathbb{E}[\sum_{0}\gamma^k r_{t+k} | s_t = s, a_t = a, \pi])
\end{equation}

A vast variety of recent advanced robot applications have been accomplished using deep reinforcement learning to teach agent complex activities including cube play \cite{akkaya2019solving}, ambidextrous robot gripping \cite{Wang2020}, categorized objects \cite{Moreira2020} and cleaning table task \cite{cruz2016learning}. For instance, Cruz et al. \cite{cruz2016learning} used an associative neural architecture to learn the available action possibilities of agents with the objects in the current context. Levine et al. \cite{levine2018learning} proposed a learning-based approach to hand-eye coordination for robotic grasping from monocular images using a large Convolution Neural Network (CNN) to learn the way to grasp objects.

\subsection{Reinforcement learning with interactive feedback}

\begin{figure*}[htbp]
    \centering
    \includegraphics[width=0.8\textwidth]{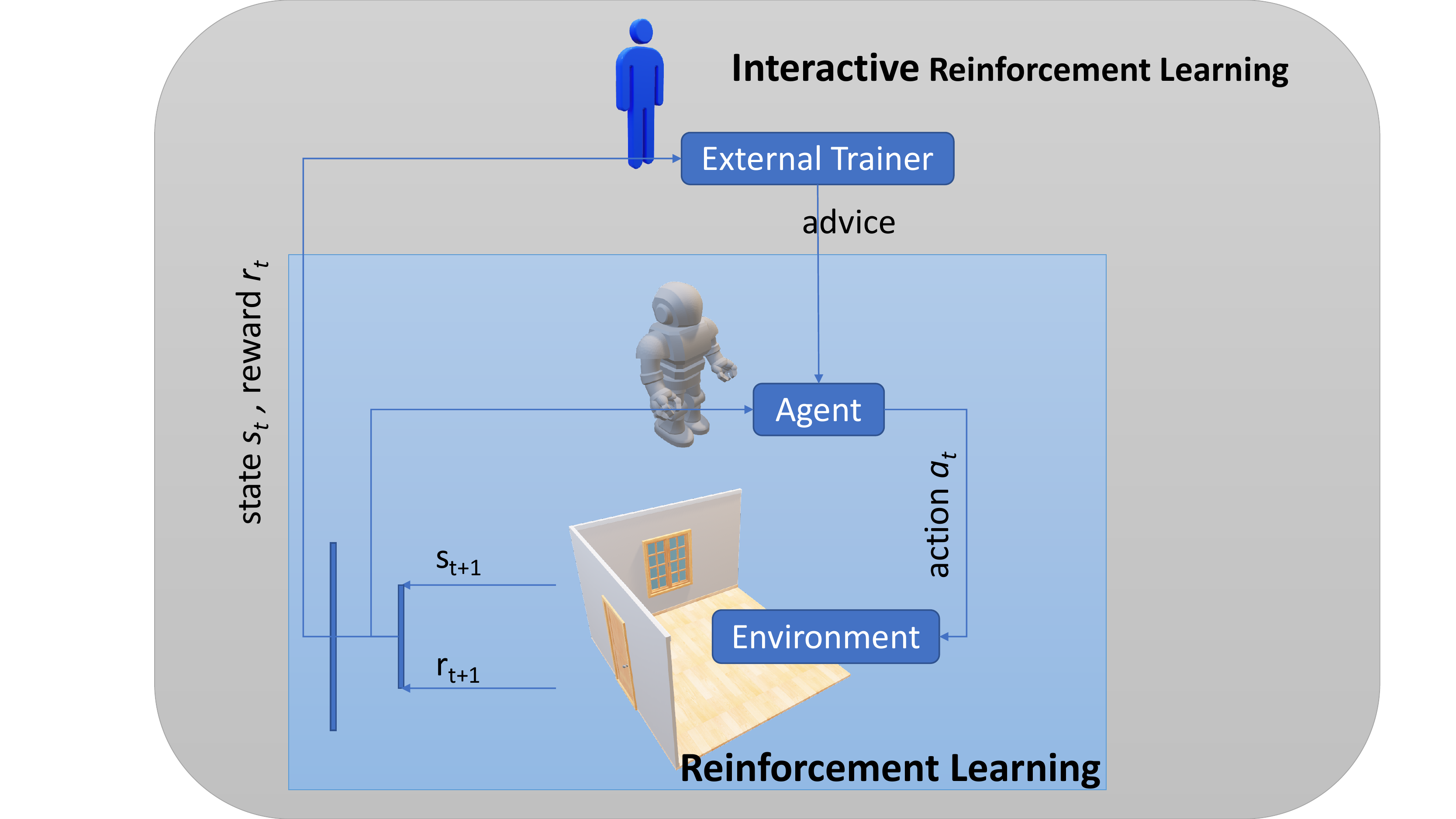}
    \caption{At state $s_t$, the agent performs action $a_t$ and obtaining reward $r_{t+1}$ and reaching the next state $s_{t+1}$ in conventional \acrshort{rl}. In \acrshort{irl} model, the involvement of an external trainer gives the agent more options to chose which action to perform next in the following iteration.}
    \label{fig:21}
\end{figure*}

In \acrfull{irl}, there is an external trainer involved in the agent’s learning process \cite{Cruz2016}. Figure \ref{fig:21} depicts the \acrshort{irl} solution, which includes a advisor who observes the learning process and offers guidance on the way to improve decision-making \cite{knox2009interactively}. The advisor can be an expert human, or an artificial agent. 

Adaptive agent behaviour is needed in domestic environments. \acrshort{irl} enables a parent-like tutor to facilitate learning by providing useful guidance in some particular situation, allowing the apprenticeship process to be accelerated \cite{bignold2021conceptual}. In contrast to an agent exploring completely autonomously, this makes for a smaller search space and hence quicker learning of the mission \cite{cruz2017agent}.

When operating alone, the next step is chosen by selecting the better known action at the current time, defined by the highest state-action pair. While \acrshort{irl} accelerates the learning process by incorporating additional guidance into the apprenticeship loop. Using \acrshort{irl}, a trainer with prior experience of the target goal is required \cite{thomaz2005real}.

There is a difference between the two main methods dedicated to feedback learning: reward shaping and policy shaping. While in the reward shaping, external trainers can assess the quality of the actions performed by the \acrshort{rl} agent, as good or bad \cite{thomaz2005real}. Using policy shaping, the actions proposed by the \acrshort{rl} agent can be replaced by more appropriate actions selected by the external trainer before implementation \cite{cederborg2015policy}.

An open problem that can significantly affect the agent’s performance is inaccurate advice from the trainer \cite{Cruz2016}, since lack of accuracy and repetitive mistakes will result in a longer training time. Human advice, on the other hand, is not 100\% correct \cite{bignold2020human}. When an advisor gives so much guidance, the agent will have limited experience in exploration because the trainer makes almost all of the decisions \cite{Matthew2014}. To address the problem, a prior study \cite{bignold2021persistent} applied to the agent a strategy of discarding or refusing advice after an amount of time, endowing an agent with the ability to work with potentially incorrect information.

%% file: components/03_approach.tex
\section{A broad-persistent advising approach}
In this section, we give more details about the proposed  \acrfull{bpa} approach that includes a generalisation model along with a persistent approach. These details are described next.

\subsection{Persistent advice}
A recent study \cite{bignold2021persistent} suggests a permanent agent that records each interaction and the circumstances around particular states. The actions are re-picked when the conditions are met again in the future. As a consequence, the recommendations from the advisor are used more effectively, and the agent's performance improves. Furthermore, as the training step is no need to provide advice for each repeated state, less interaction with the advisor is required. However, in this experiment, we limit the research to keeping the same number of interactions to the trainer to investigate the effectiveness of \acrshort{bpa} approach in continuous domain.

As aforementioned, there is an issue relating to inaccurate advice. After a certain amount of time, a mechanism for discarding or ignoring advice is needed. \acrfull{ppr} is a strategy for improving \acrshort{rl} agents that use advice \cite{fernandez2006probabilistic}. Where various exploration policies are available, \acrshort{ppr} uses probabilistic bias to decide which one to choose, with the intention of balancing between random exploration, the use of a guideline policy, and the use of the existing policy.

\begin{figure}[htbp]
    \centering
    \includegraphics[width=0.48\textwidth]{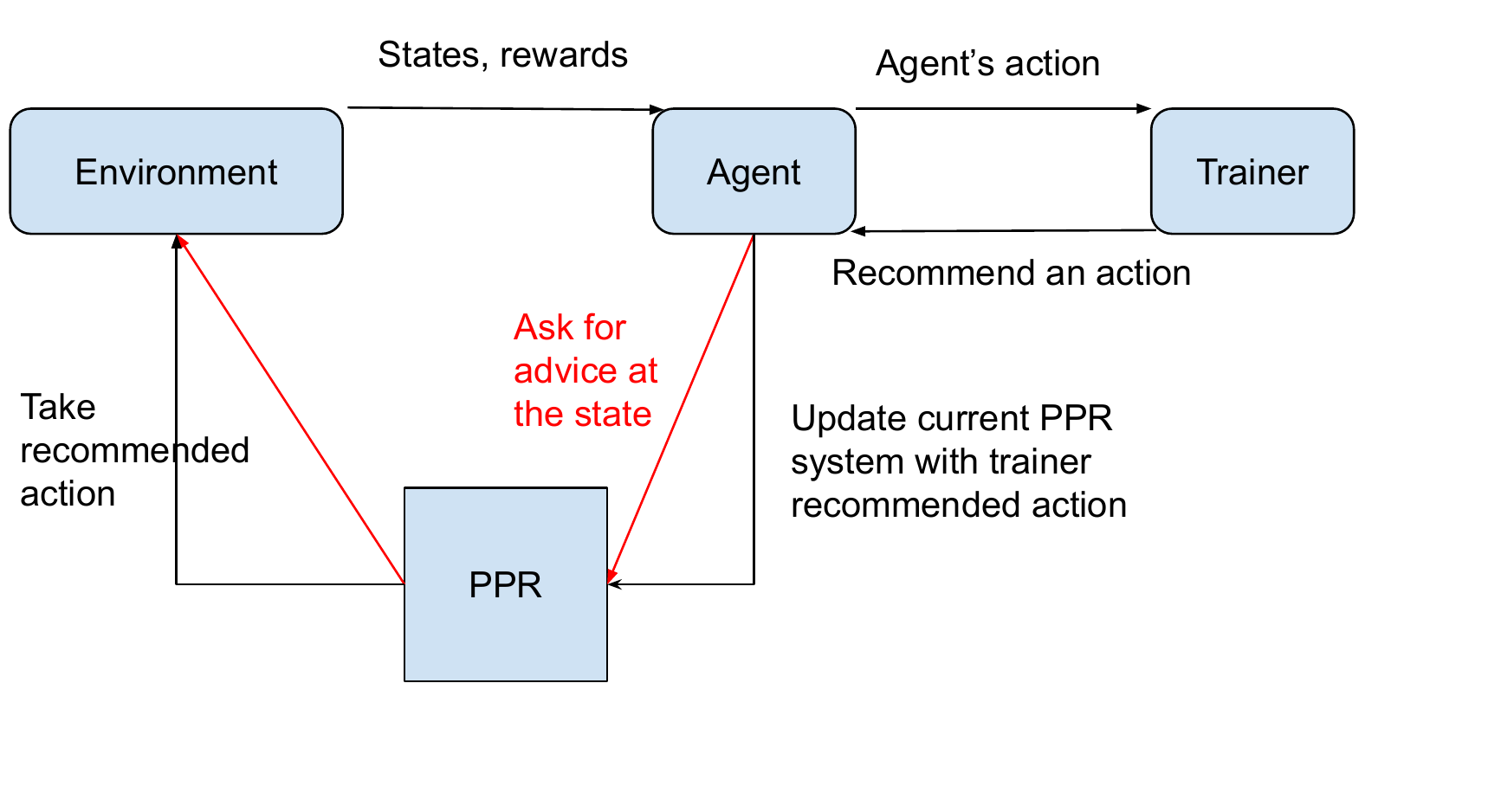}
    \caption{Process flow of an interactive reinforcement learning agent using \acrshort{ppr} system. After receiving advice from the trainer, the agent will store the action into the \acrshort{ppr} model. In the iteration not receiving advice from the trainer, the agent will check and reuse the old advice (red arrows) from the past.}
    \label{fig:31}
\end{figure}

Figure \ref{fig:31} denotes an example of \acrshort{irl} using \acrshort{ppr}. The advising user has the opportunity to engage with the agent at each time point. When there is an interaction, the model is updated. At the time advice is firstly recommended, it is assumed that the agent will carry it out the suggested action, regardless of the setting of \acrshort{ppr}. \acrshort{ppr} is used the time step when the agent did not receive advice from the trainer, which flow is denoted by red arrows. First, the agent's policy is examined to see whether any advice is applicable to the existing state. If the current policy suggests an action, the action is taken with the determined by the \acrshort{ppr} selection policy.

\acrshort{ppr} is used where an agent chooses an action in a time step where the user has not recommended a prior action, which is denoted by red arrows. First, the agent's policy is examined to see whether any advice is applicable to the existing state. If the current policy suggests a action, the action is taken with the determined by the \acrshort{ppr} selection policy.

\subsection{Broad advice}

To use PPR, we need a system to store the used pairs of state-action. When the agent arrives at a certain state at a time step, agents using PPR need to check with the system if this state has been suggested by the trainer in the past. If there is advice in the memory of the model, the agent can use the option to reuse the action. However, there is a problem when using PPR in infinite domains. We cannot build a system that stores state-action pairs with infinite state values. In addition, when the amount of state becomes too large in space, which is equivalent to infinity, the possibility that agents revisit exactly the same state will be very small. Therefore, building this model will become cumbersome and inefficient in large spaces. 

\begin{figure}[htbp]
    \centering 
    \includegraphics[width=0.48\textwidth]{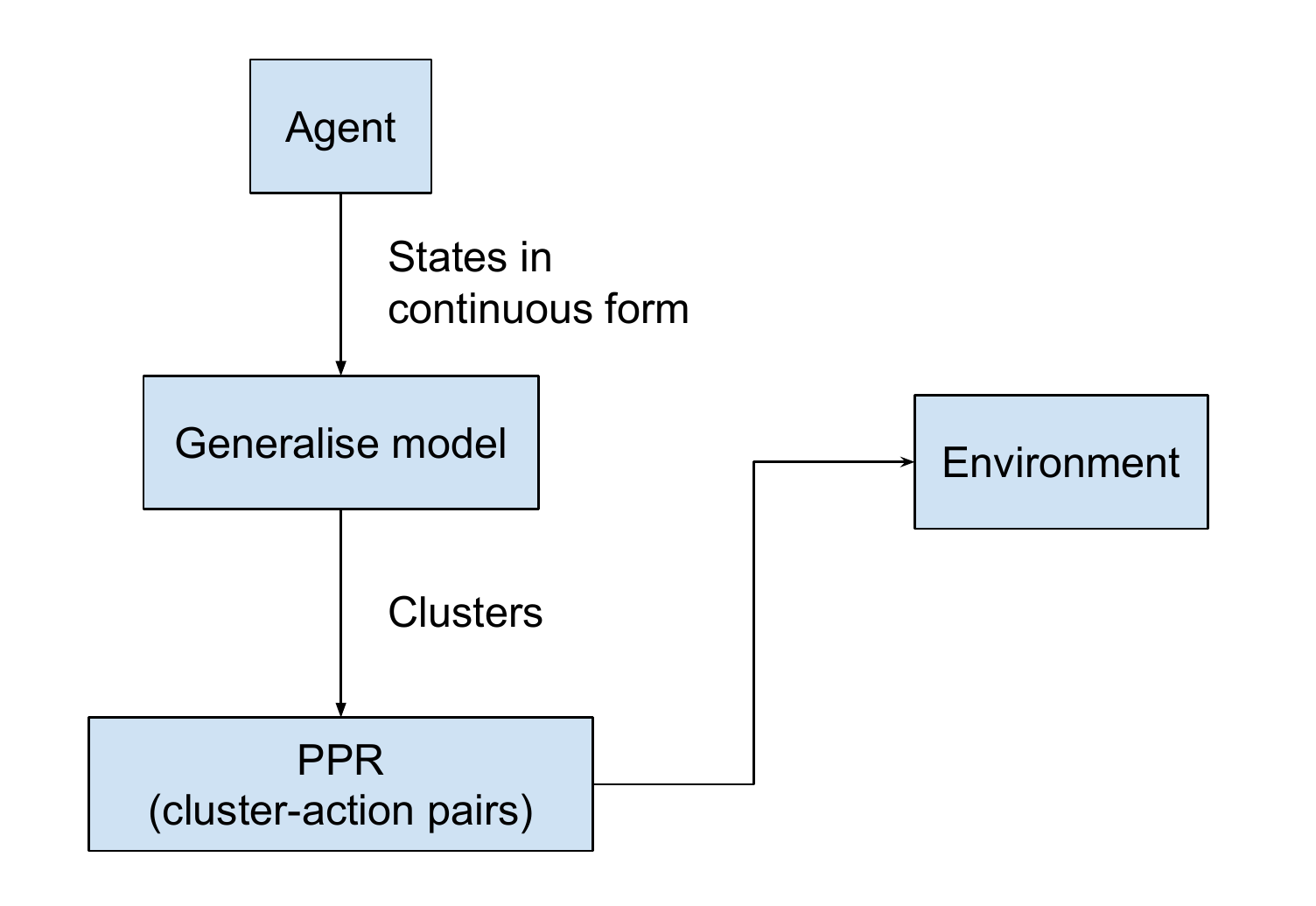}
    \caption{Broad advice transform continuous states into finite clusters. Hence, the state-action pair becomes cluster-action used in the \acrshort{ppr} model.}
    \label{fig:32}
\end{figure}

\acrshort{bpa} includes a model for clustering states and then building a system for cluster-action pairs instead of traditional state and action pairs. The proposed model is shown in Figure \ref{fig:32}. When the agent receives current state information from the environment and it does not receive any advice from the trainer, the agent will use PPR by injecting the state into the generalisation model and defining its cluster. Then proceed to consider whether any advice pertains to the current cluster. If there is an action recommend in the past, the agent can reuse it with the PPR selection probability, or use default action as $\epsilon$-greedy.

The generalisation model we use in this paper is the $k$-means algorithm. $k$-means is one of the most popular clustering methods \cite{park2013novel}. $k$-means is simple to implement, and its complexity scales well with a higher number of data. However, the user must decide on the number of clusters beforehand \cite{madhulatha2012overview}. We used the elbow technique to specify the number of clusters \cite{humaira2020determining}. It is the visual graphic approach that was generated from the Sum Square Error (SSE) computation. This technique is based on the idea that the number of clusters should be chosen so that adding another cluster does not cause significantly improved modeling. The early clusters will provide a lot of information, but at some point, the marginal gain will drop drastically, giving the graph an angle. At this angle, the correct $k$-number of clusters is determined, thus called "elbow criteria".

%% file: components/04_experimental.tex
\section{Experimental environments}

\subsection{Cart pole gym environment}
The deep reinforcement learning environment is implemented using the well-known library of AI gym environments \cite{AIgym}. First, we build the Cart Pole environment. In this environment, there is a pole that is attached to the cart. The carriage can move by applying force to the left or right. The purpose of this problem is to prolong the time while avoiding the pole falling down. The terminal condition is that the pole deviates more than 15 degrees from the vertical or the wagon moves 2.4 units from the center. The cart pole \acrshort{mdp} is defined as follow: 

\begin{itemize}
  \item State: The state vector has a continuous representation with four attribute which represent for cart position, cart velocity, pole angle and pole velocity.
  \item Action: The cart can perform two actions on the track: go to left or right.
  \item Reward function: As long as the agent holds the pole in a vertical position, a reward equal to 1 is awarded, and if it drops, or goes beyond the boundaries of the track, the reward is equal to 0.
\end{itemize}

\begin{figure}[htbp]
    \centering
    \includegraphics[width=0.4\textwidth]{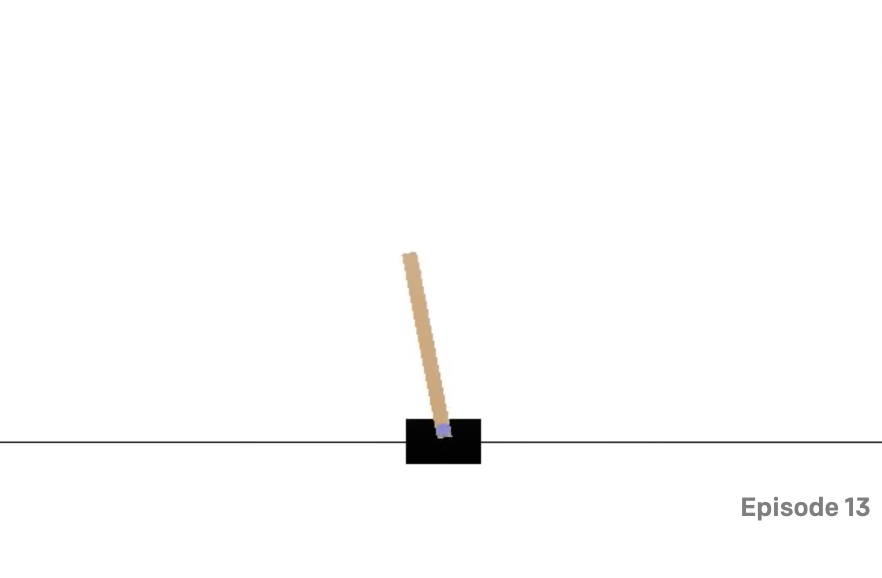}
    \caption{A graphical representation of the Cart Pole environment. The goal is to keep the pole balanced while applying forces to the carriage. The terminal condition is that the pole deviates more than 15 degrees from the vertical or the wagon moves 2.4 units from the center.}
    \label{fig:41}
\end{figure}

Figure \ref{fig:41} below denotes a graphic of the Cart Pole in the AI-gym environment.

\subsection{Domestic robot environment}
\begin{figure}[htbp]
    \centering
    \includegraphics[width=0.4\textwidth]{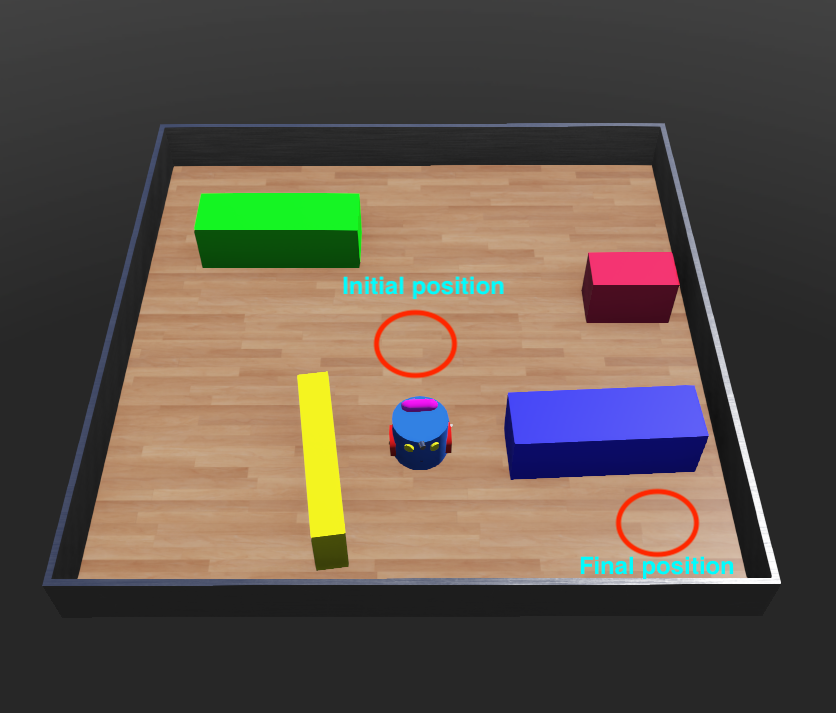}
    \caption{A example of Webots environment with initial position and final position. The robot has goals to go from initial position to final position while avoiding obstacles. The robot will be returned to its initial position after any collision. }
    \label{fig:42}
\end{figure}

Additionally, we also build an environment for domestic robots using Webots, given the overall good performance shown previously \cite{ayala2020comparison}. In this environment, the goal is to train the robot to go from the initial position to the target position. Figure \ref{fig:42} denotes a graphic of our experimental environment in Webots.

The robot is equipped with distance sensors on its left and right eyes. The robot is completely unaware of its current position in the environment. The robot can only choose one of three actions: go straight at 3m/s, turn left, or turn right. At each step, the robot will be deducted 0.1 points if it uses the action of turning left or right, no points will be deducted if it chooses to go straight. This is to optimize the robot's straight movement and avoid the robot running in circles by turning left or right continuously. The robot is equipped with a few touch sensors next to it, to detect the collision with the environment. The robot will be returned to its initial position and receive 100 penalty points every time it collides on the way. The robot does not know where the touch sensor is located relative to itself, the only information it receives is whether it is a collision with obstacles or not. When the robot goes to the finish position located in the lower right corner of the environment, the robot is considered to complete the task and be rewarded with 1000 points.

\begin{figure}[htbp]
    \centering
    \includegraphics[width=0.4\textwidth]{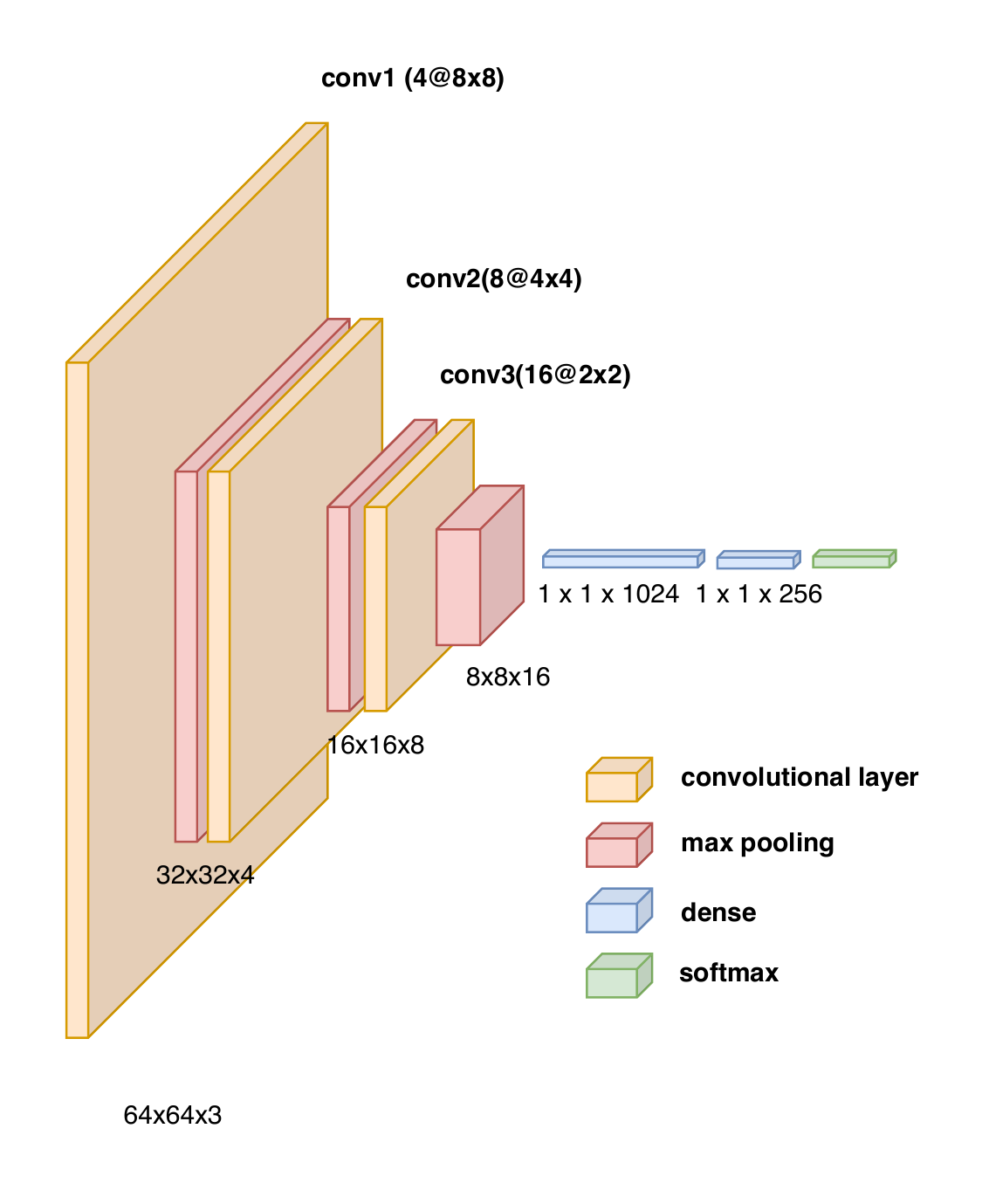}
    \caption{\acrshort{cnn} architecture with 64x64 RGB image as input, group of three convolution layers, three max-pooling layers, two dense full connected layers, and a softmax function at the output}
    \label{fig:cartpole}
\end{figure}

To decide on the next action the robot will choose, the robot's supervisor will use the image taken from the top of the environment to enter the \acrfull{cnn} system to decide. The \acrshort{cnn} system built in this environment will be a system whose input is 64x64 image RGB channels. This architecture is inspired by similar networks used in other \acrshort{drl} works \cite{Moreira2020,krizhevsky2012imagenet}. In more detail, we use 4 kernels with size 8x8. The second layer is 8 kernels with size 4x4, and the last layer convolution is 16 kernels with size 2x2. Following each convolution network layer is a 2x2 max-pooling layer. Finally, there is a flatten and dense layer with 256 neurons fully connected with the output layer. The network architecture is described in Figure \ref{fig:cartpole}.

The environment \acrshort{mdp} is defined as follow: 

\begin{itemize}
  \item State: RGB image size 64x64 taken from the top of the environment.
  \item Action: Three actions: go straight at 3m/s, turn left, or turn right
  \item Reward function: Turn left, right: -0.1; Go straight: 0; Collision: -100; Reach to final position: 1000
\end{itemize}

\subsection{Interactive feedback}

While the interactive agent's human-related approach to learning is one of its greatest strengths, it may also be its greatest weakness \cite{Bignold20211, Cruz20181}. Advice with good accuracy given in the proper time will help the agent a lot in speeding up the speed of finding the optimal solution. However, in the case when the agent only gives advice with low accuracy and in high frequency, that not only does not help the agent but also brings it to a dead road and is much more time-consuming than no interaction situation. Furthermore, human experiments are costly, time-consuming, have problems of repeatability, and can be difficult to recruit volunteers. Therefore, during the early stages of the agent, we suggested that simulating human interactions would be much more convenient. 

To compare agent performance, information about interactions, agent steps, rewards, and interactions are recorded. To identify the efficiency of \acrshort{bpa}, we need to test the experiment with three cases: No interactive action, interactive actions without \acrshort{bpa}, and interactive actions with \acrshort{bpa}.

In addition, each use case of the simulated user will have different advice's accuracy and frequency. Frequency is the availability of the interaction of the advisor at the given time step. The higher frequency, the advisor has more rate for giving advice to the agent. Accuracy is a measure of the precision of advice provided by an advisor. When the advisor's accuracy is high, the action would be proposed precisely as the advisor's knowledge. On the contrary, the action proposed is different from the advisor's knowledge. The frequency and accuracy of the real-world simulation were simulated using data from a human test \cite{bignold2020human} and the article \cite{bignold2021persistent} which is described in Table \ref{table:three_simulated_advisor}. The value of frequency and accuracy of advice is beyond the scope of this study.

Each use case of the simulated user will have different advice's accuracy and frequency. Accuracy is a measure of the precision of advice provided by an advisor. When the advisor's precision is high, the action would be proposed precisely as the advisor's knowledge of the environment. On the contrary, the advisor would propose not optimal action based on how it knows about the environment. Frequency is the availability of the interaction of the advisor at the given time step. The higher frequency, the advisor has more rate for giving advice to the agent. Accuracy is a measure of the precision of advice provided by an advisor. When the advisor's accuracy is high, the action would be proposed precisely as the advisor's knowledge. On the contrary, the action proposed is different from the advisor's knowledge. Accuracy and frequency of three kinds of agents are used with value described in Table \ref{table:three_simulated_advisor}. Optimistic simulated agents have 100\% accurate advice and always provide advice on every time step. Realistic simulated agents use accuracy and frequency value from results in a human trial \cite{bignold2020human, bignold2021persistent}. The pessimistic value of frequency is 0\%, however, it works the same as in the case without interactive feedback. Therefore, we use half of the realistic value for the case with the least interaction of the advisor. The accuracy and frequency value advice is beyond the scope of this study.

\begin{table}[ht]
    \begin{center}
    \caption{The three simulated users designed for the experiments. These users will are not intended to be compared against each other,  rather than comparing with persistent counterpart}
        \begin{tabular}{| m{10em} |  m{7em} |  m{7em} |} 
            \hline
            \textbf{Agent} & 
            \textbf{Frequency} & 
            \textbf{Accuracy} 
            \\ 
            \hline
            Pessimistic Advisor  & 
            23.658\% & 
            47.435\% 
            \\ 
            \hline
            Realistic Advisor  & 
            47.316\% & 
            94.87\% 
            \\ 
            \hline
            Optimistic Advisor & 
            100\% & 
            100\%
            \\ 
            \hline
        \end{tabular} \label{table:three_simulated_advisor}
    \end{center}

\end{table}

The corresponding frequency and accuracy values above are pessimistic value, realistic value and optimistic value, respectively. The pessimistic value of frequency is 0\%, however it works the same as in the case without interactive feedback. Therefore, we use half of the realistic value as the case with the least interaction of the advisor.
    
Experiments are performed for each case and the indicator of how accumulated reward can achieve the optimal policy will be recorded to compare the result between many approaches. The more reward the agent takes, the better result of the method is.

\subsection{Generalise model and probabilistic policy reuse}
\begin{figure}[ht]
    \centering
    \includegraphics[width=0.48\textwidth]{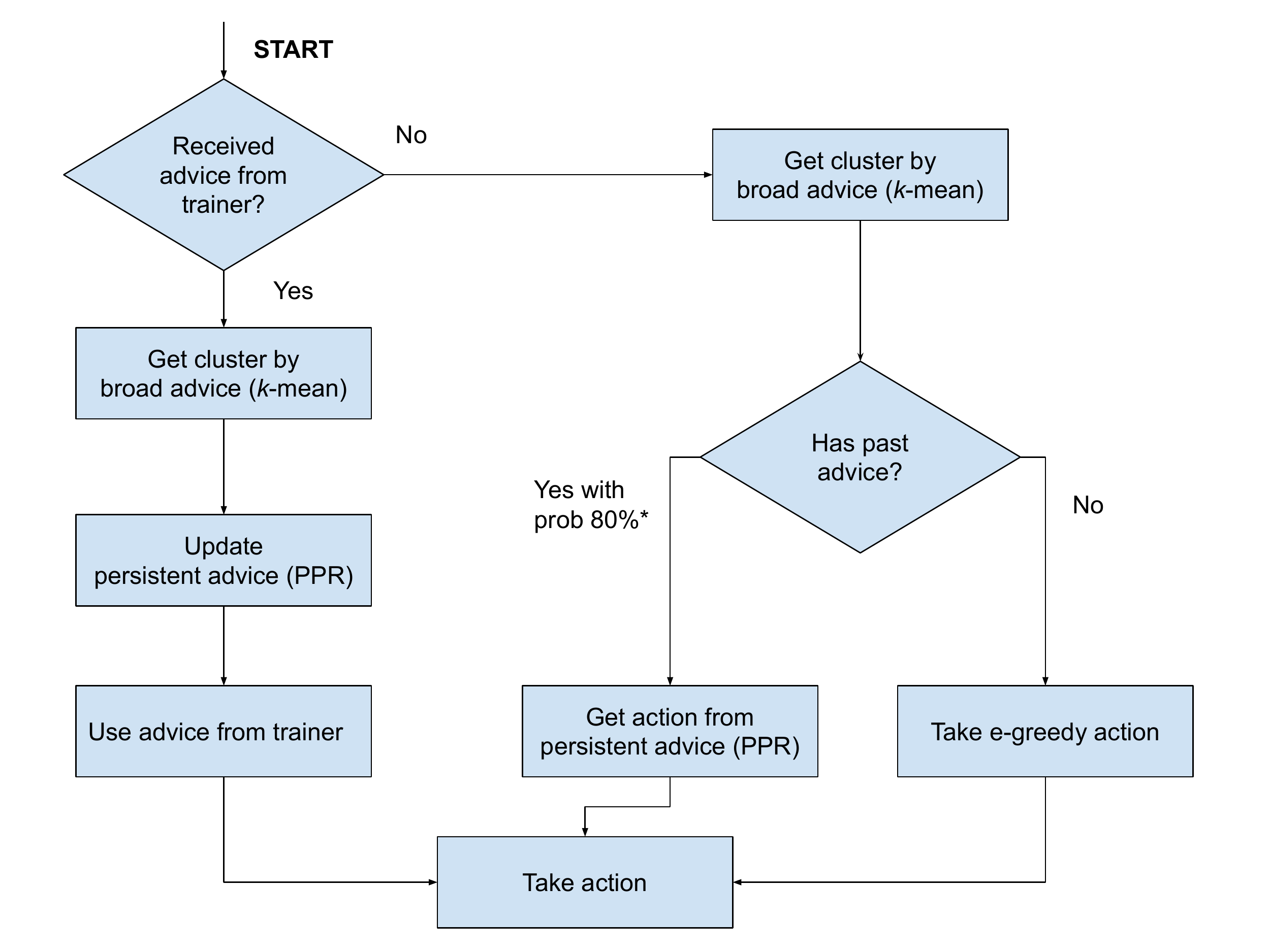}
    \caption{Flow of using broad-persistent advising. The agent will reuse previously obtained advice with 80\% chance (decays over time) and perform its exploration policy for the remaining change (20\%).}
    \label{fig:43}
\end{figure}

Next, we demonstrate the use of and broad advice and persistent advice using \acrfull{ppr}. The flow of using PPR is depicted as shown in Figure \ref{fig:43}. Initially, the agent reuses the action using PPR with a certain chance if the current state has been recommended by the trainer in the past. At the current work, we use chance value at 80\% which is also used in previous research \cite{bignold2021persistent}. This probability decreases by 5\% for each step. With the remaining 20\%, the greedy action policy is selected. 

Algorithm \ref{alg:bpa} shows the process flow for selecting an action using \acrshort{bpa} approach to assist a learning agent. 

\begin{algorithm}[H]
\caption{Interactive reinforcement learning with a \acrshort{bpa}} \label{alg:bpa}
\begin{algorithmic}[1]
\State Built $k$-means model with states from trainer
\State Initialize environment selecting $s_t$
\ForAll{(episodes)}
    \Repeat
        \If{(have feedback)}
             \State Get recommend action $a_t$
             \State Get cluster $c_t$ by using $k$-means 
             \State Add pair ($c_t$, $a_t$) to ppr
        \Else
            \If{($rand(0,1) < \epsilon$)}
                \State Get $c_t$ by using $k$-means
                \If {($c_t$ is available in ppr)}
                    \State Get $a_t$ is reuse action from ppr
                \Else
                    \State Random action $c_t$ from environment
                \EndIf
            \Else
                \State Choose action $a_t$ using $\pi$
            \EndIf
        \EndIf
        \State Perform action $a_t$
        \State Observe next state $s_{t+1}$
    \Until{($s$ is terminal)}
    \State Update policy $\pi$
\EndFor 
\end{algorithmic}
\end{algorithm}

The model will be tested with the following agent listed below:

\begin{itemize}
    \item {
    Baseline \acrlong{rl}: The model will be trained basically and collect information from the environment without using any interactive feedback or evaluation from the trainer. It is used as a benchmark.}
    \item {
    Non-persistent \acrlong{rl}: The agent is assisted by multiple type of users with in mentioned before in Table \ref{table:three_simulated_advisor}. After taking recommendation from trainer and execute the action, the agent will discard the advice. When the agent come to the similar state again in the future, it cannot recall the previous recommendation and performs an $\epsilon$-greedy action instead.}
    \item {
    Persistent \acrlong{rl}: This agent is supported by a trainer and \acrshort{ppr} system. The trainer can suggest an action in each time step for the agent to take. If recommended, the learning agent will perform on that time step and retain the recommendation for reusing when it visits the similar state in the future. When an agent accesses similar state it has previously suggested, it will perform that action with the probability determined by the PPR action selection rate.}
\end{itemize}

%% file: components/05_results.tex
\section{Results}~\label{sec:evaluation}
\subsection{Cart pole domain}
In this section, we proceed to display the results three types of agents proposed above, including: baseline \acrshort{rl} for bench marking, non-persistent \acrshort{rl}, persistent \acrshort{rl}. For agents of the type non-persistent \acrshort{rl} and persistent \acrshort{rl} we conduct tests on different frequency and accuracy of feedback, called optimistic user, realistic user and pessimistic user. The method for all the agents are tested with the same hyper-parameters as follows: initial value of $\epsilon$ = 1, $\epsilon$ decay rate of 0.99, learning rate $\alpha$ = 0.01, and discount factor $\gamma$ = 0.99 during 500 episodes. To better display, we computed the average value of the last 100 rewards instead of the current episode reward. We inspired the idea done on this article with the same result from cart pole environment \cite{kumar2020balancing}. 

\begin{figure}[ht]
    \centering
    \begin{subfigure}[b]{0.9\linewidth}
        \includegraphics[width=\textwidth]{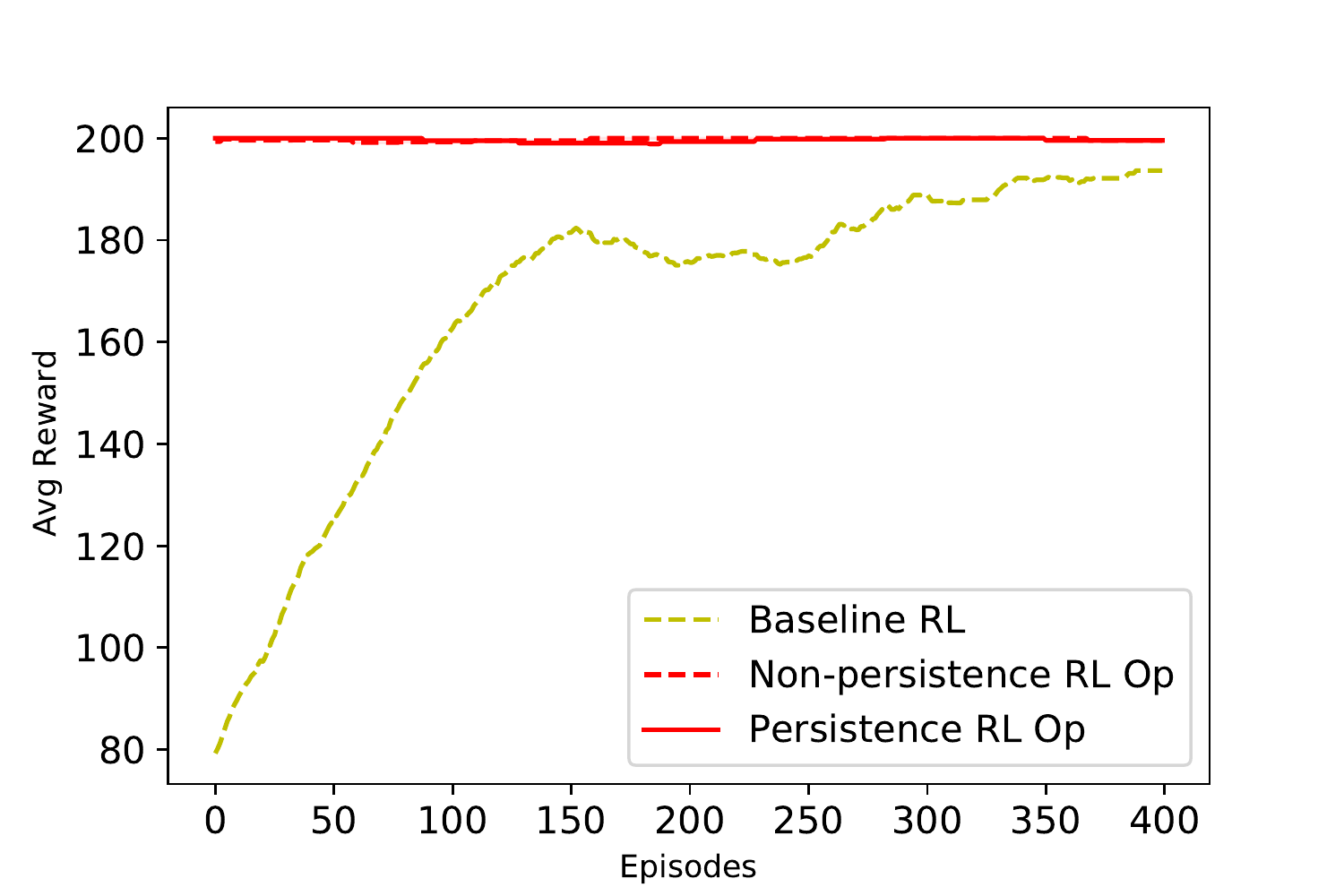}
        \caption{Optimistic agents}
        \label{fig:op}
    \end{subfigure}
    \begin{subfigure}[b]{0.9\linewidth}
        \includegraphics[width=\textwidth]{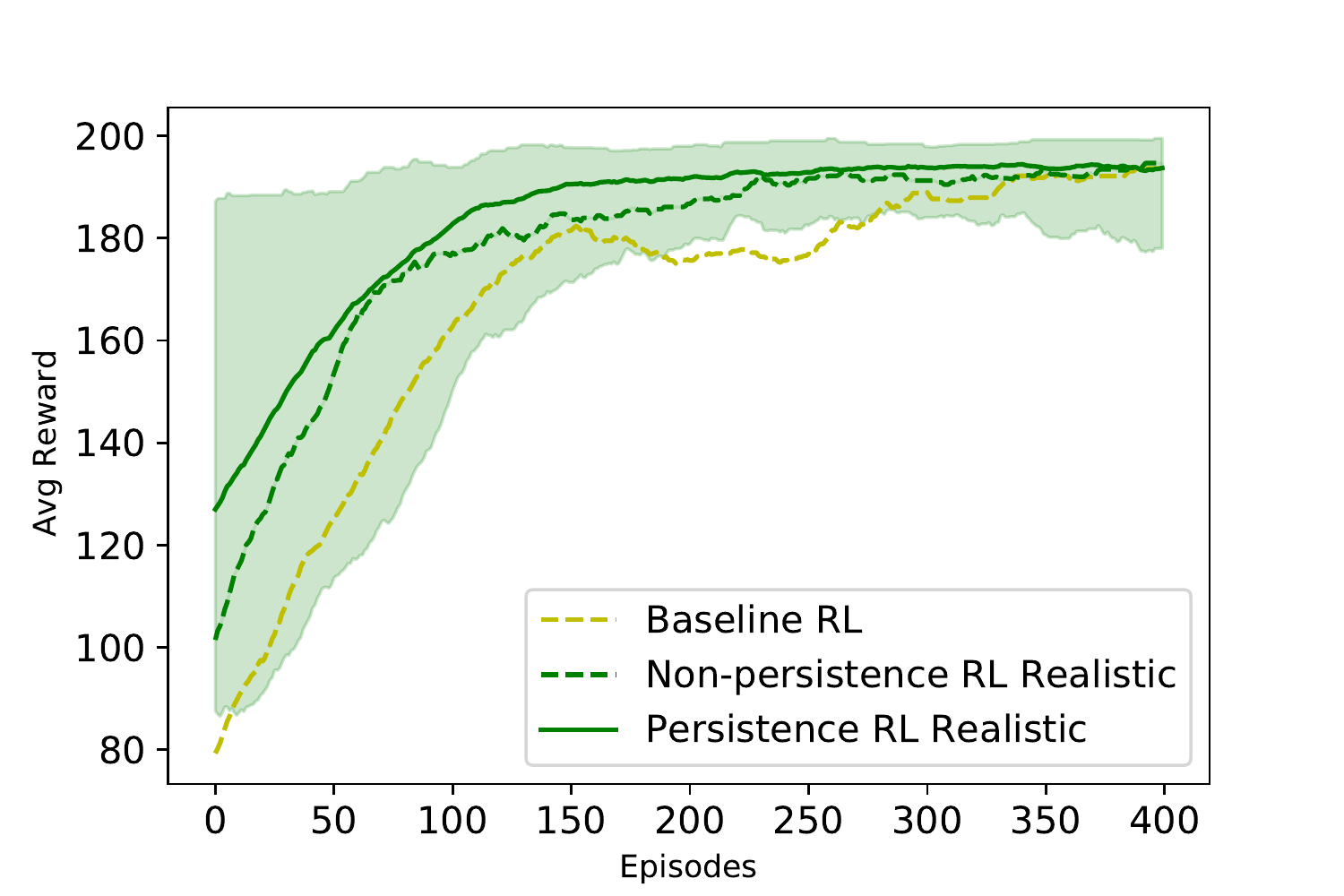}
        \caption{Realistic agents}
        \label{fig:re}
    \end{subfigure}
    \begin{subfigure}[b]{0.9\linewidth}
        \includegraphics[width=\textwidth]{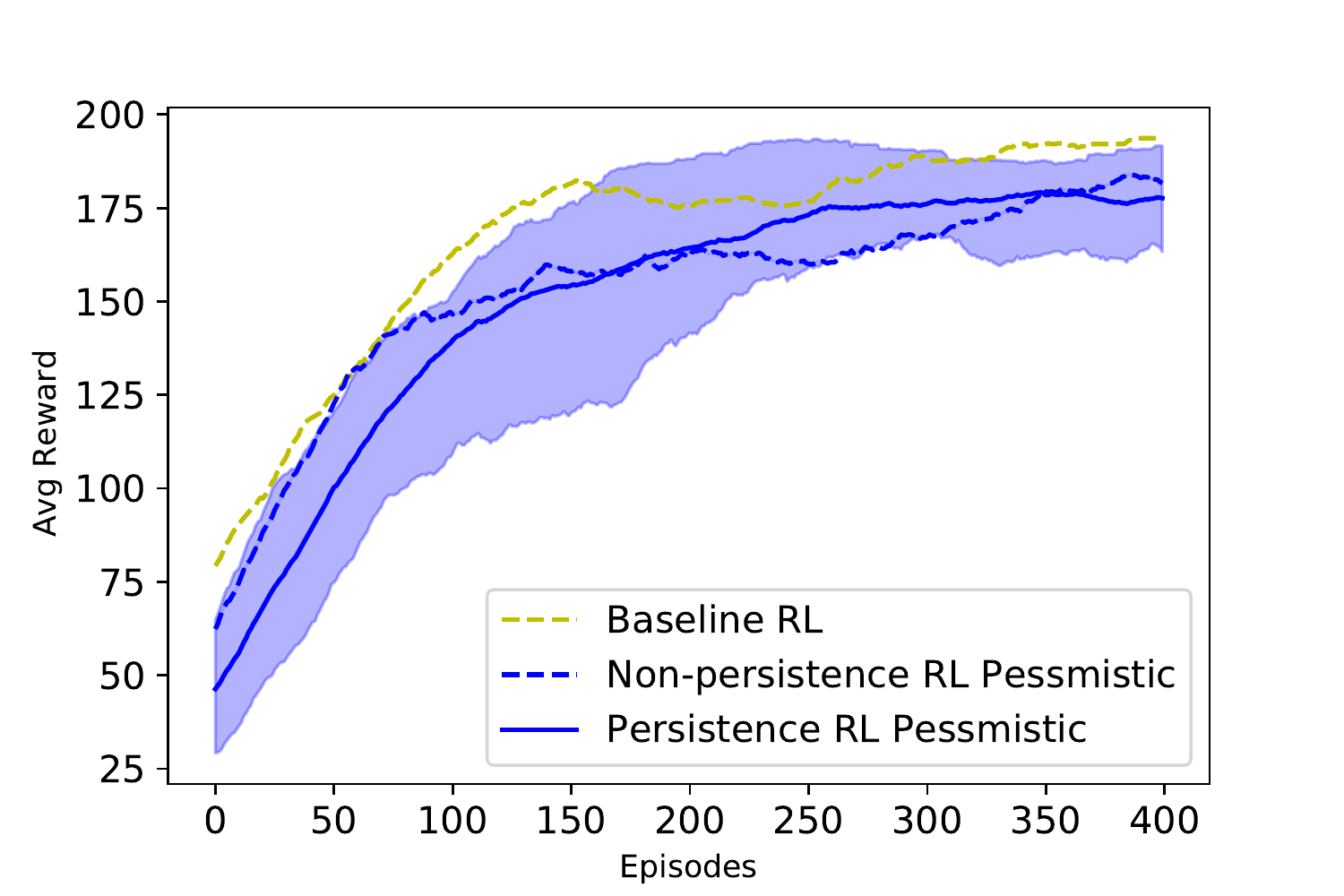}
        \caption{Pessimistic agents}
        \label{fig:pe}
    \end{subfigure}
    \caption{The comparison for persistent agents, non-persistent agents, and baseline each kind of agent in deep reinforcement learning built with cart pole environment. The shaded area indicates the deviation between the minimum and maximum values of the agent value after multiple training.}
    \label{fig:51}
\end{figure}

The results obtained are shown in Figure \ref{fig:51}. Optimistic, realistic, and pessimistic agents are run for five times and are represented by red, green, and blue lines respectively. The shaded area indicates the standard deviation of the agent's reward after multiple training. Overall, all interactive agents outperformed the autonomous one (baseline \acrshort{rl} in yellow), except the pessimistic agents. Agents which receive advice from the instructor make fewer mistakes, especially in the early stages of the learning process, and can learn the task in fewer episodes. However, in this work we want to compare pairs of non-persistent and persistent agents with the same style delivered advice to verify if
the \acrshort{bpa} approach implementation is indeed effective. 

The agents assisted by optimistic achieved the maximum score of \acrshort{dirl} algorithms at very early after a few episodes. Because the trainer always makes decisions for the agent (100\%), and this decision is absolutely correct (100\%). In this experiment, the agents did not even have any chance to make their own decisions or use PPR, the trainer made all decision.

On the contrary, agents supported by pessimistic users have different results, but in fact, neither of them can solve the problem. On many runs in both non-persistent and persistent cases, the agent failed to achieve convergence. Both cases are considered worse than the baseline. This can be explained because the accuracy of advice for pessimistic agents is only 23.658\%

On the graph of the agents being helped by realistic trainers, we can see that using PPR produces slightly better results than the non-using counterpart. Persistent agents not only have a better initial reward but also can achieve convergence results 100 episodes earlier than non-persistent agents. This difference in learning rates is due to the fact that the agent retains and reuses advice. In this experiment, the realistic agent has a 47.3\% chance to interact with the trainer and the agent will withhold or not withhold the advice from the trainer depending on whether it is a persistent agent or non-persistent agent. However, the persistent agent will retain and reuse the advice with an 80\% probability (decreasing over time) for any state in which it has received the advice in the past. As long as the stored advice is accurate enough, the persistent agent will learn faster because they use the advice more often.

In this experiment, we focused on implementing persistent advice in a continuous environment. The ratio of the number of interactions with our trainers remains the same: for example 47.316\% with the realistic agent. When receiving advice from the trainer, the agent will always prioritise executing this recommended action. Therefore, the number of interactions using the BPA method is equivalent to not using it. Table \ref{table:feedback-iteration} shows the average number and percentage of interactions that occurred for each agent. Both non-persistent and persistent agents use the interaction rate according to Table \ref{table:three_simulated_advisor}. We can see that the number of interactions of every pair of optimistic, realistic and pessimistic agents are similar in the experiment.

\begin{table}[ht]
    \begin{center}
    \caption{The average number of interactions in experiment for each kind of agent, and the percent compare with total steps taken}
        \begin{tabular}{lll}
        \toprule
        \textbf{Agent} & \multicolumn{2}{c}{\textbf{Interaction}} \\
        \cmidrule(r){2-3}
        & \textbf{Non-persistent} & \textbf{Persistent} \\
        \midrule
        Optimistic Advisor  &   99796 (100\%)    &   99846 (100\%) \\
        Realistic Advisor   &   40976 (47.15\%)  &   41832 (47.1\%) \\
        Pessimistic Advisor &   18034 (23.62\%)  &   16685 (23.76\%) \\
        \bottomrule
        \end{tabular}     \label{table:feedback-iteration}
    \end{center}
\end{table}

\begin{figure}[ht]
    \centering
    \begin{subfigure}[b]{0.9\linewidth}
        \includegraphics[width=\textwidth]{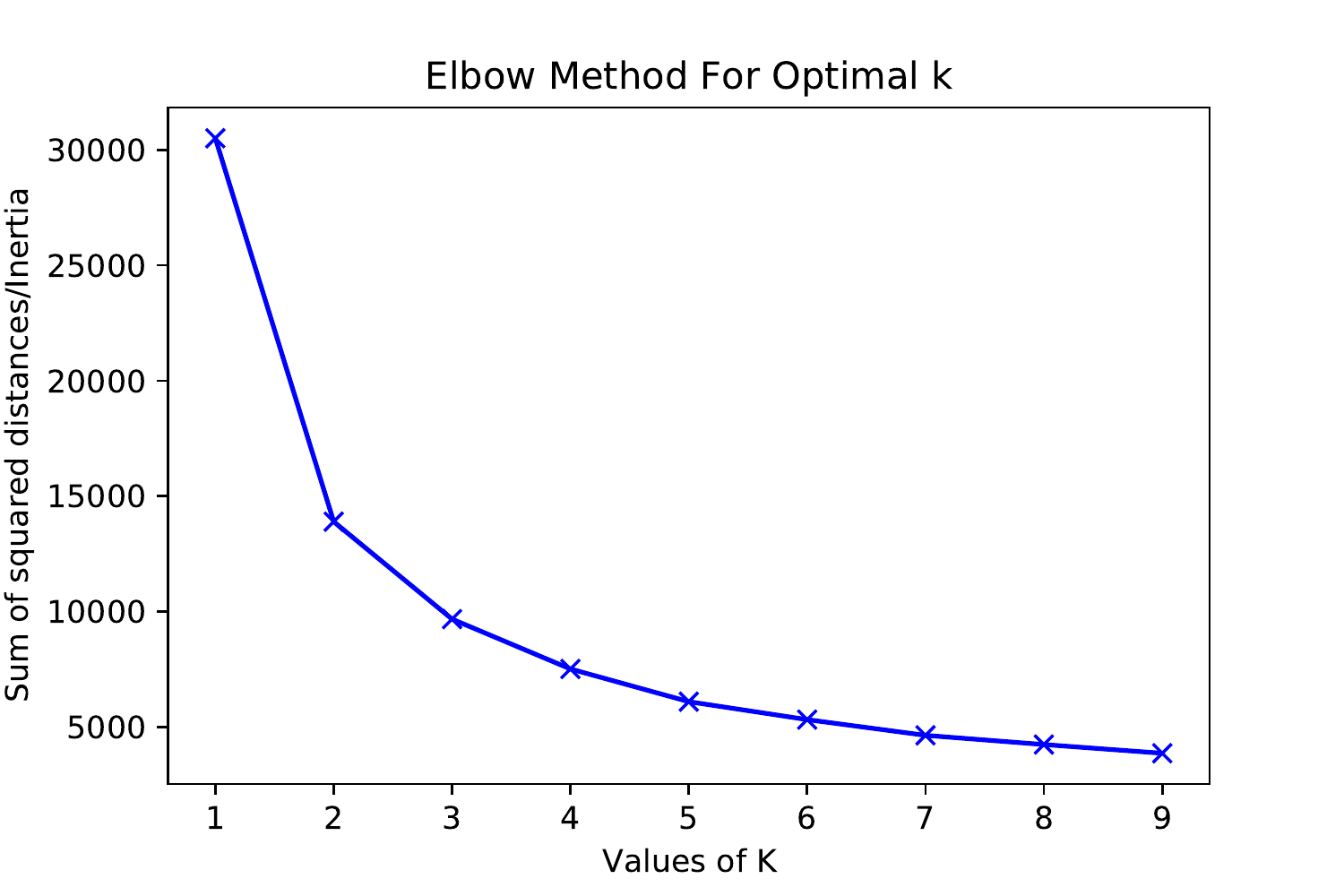}
        \caption{Elbow method}
        \label{fig:52a}
    \end{subfigure}
    \begin{subfigure}[b]{0.9\linewidth}
        \includegraphics[width=\textwidth]{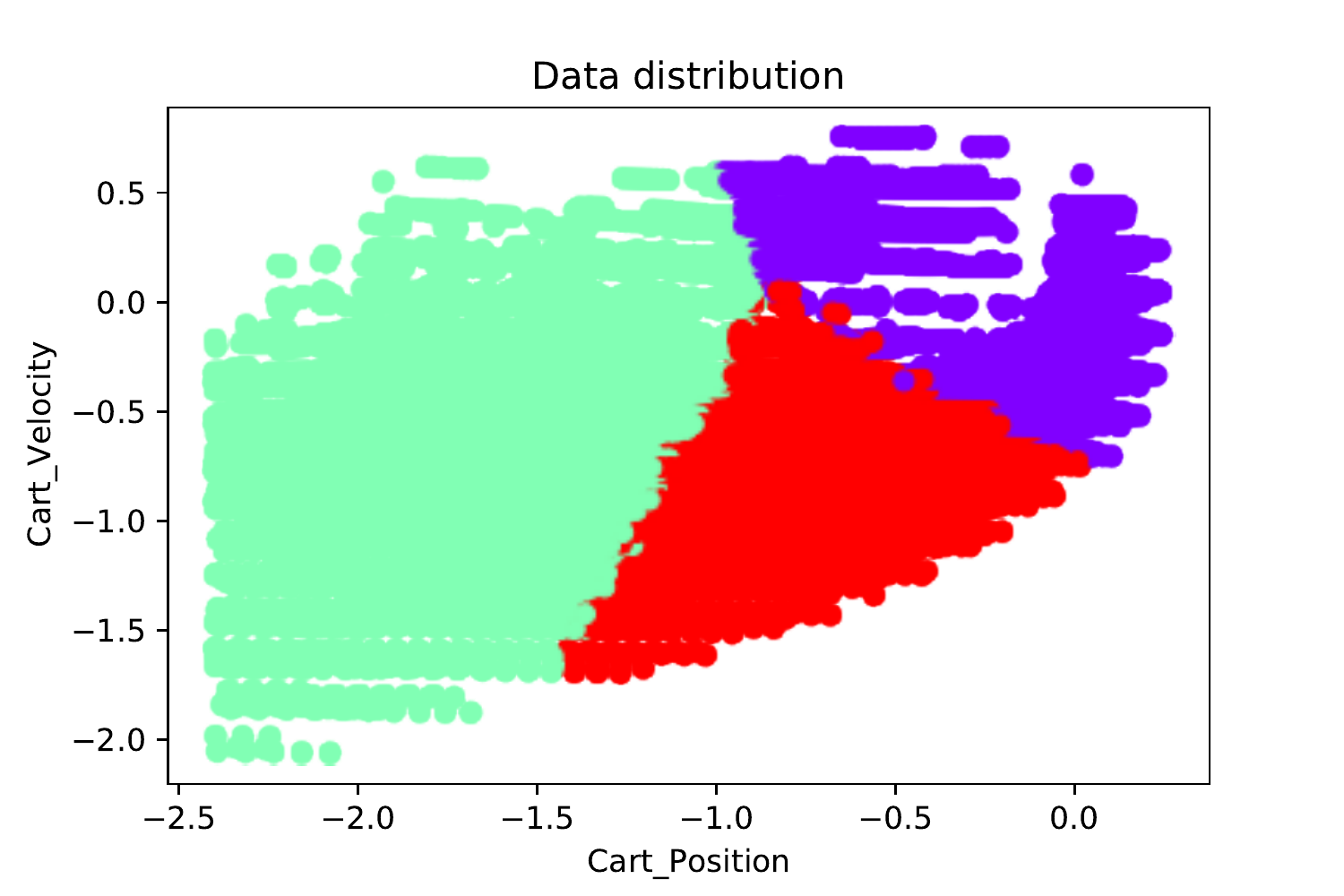}
        \caption{Distribution}
        \label{fig:52b}
    \end{subfigure}
    \caption{Total of squared distance for value k from 1-9 and distribution for 50000 states with value k = 3 at cart position and cart velocity attributes}
    \label{fig:52}
\end{figure}

Figure \ref{fig:52a} shows a graph using the elbow method to specify the number of clusters as a parameter of $k$-means, using the data of 50000 states, the same number as the experiment of cart pole, in the actual running environment. The elbow method shows the best value for using k at the value 3. Figure \ref{fig:52b} displays the data distribution in cart position and cart velocity attributes axes at the value k = 3.  

\subsection{Webots domain}

In this scenario, we focus only to examine the results for the realistic agent, because this can be transferred to the real-world scenarios in a more rational manner. The method is tested with the following hyperparameters: initial value of $\epsilon$ = 1, $\epsilon$ decay rate of 0.99, learning rate $\alpha$ = 0.01, and discount factor $\gamma$ = 0.99 during 500 episodes. We use average value of the last 100 rewards instead of the current reward only.

\begin{figure}[ht]
    \centering
    \includegraphics[width=0.95\linewidth]{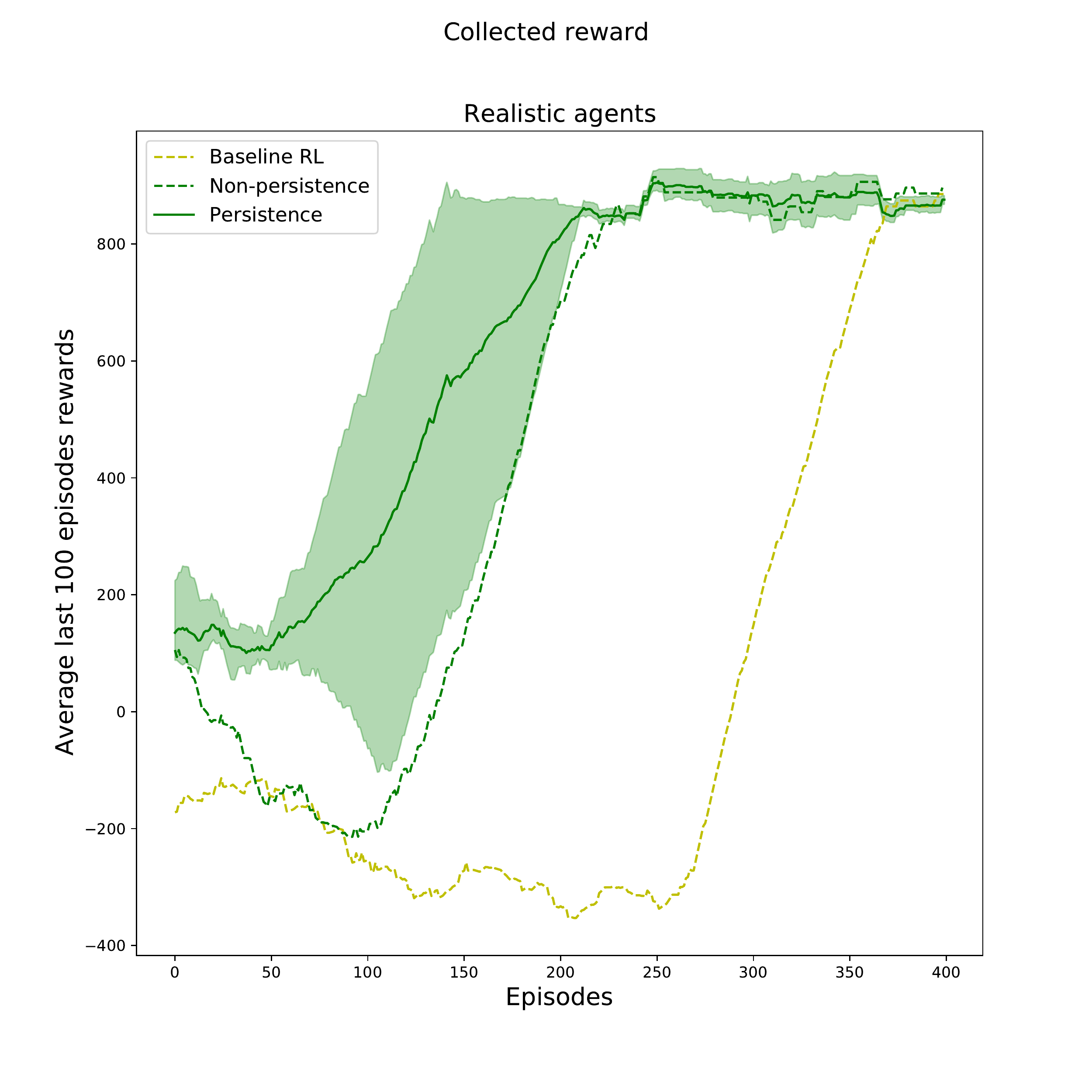}
    \caption{Result for deep reinforcement learning with autonomous agent, non-persistent agent and persistent agent built with Webots domestic robot environment.}
    \label{fig:53}
\end{figure}

The results obtained are shown in Figure \ref{fig:53}. Non-persistent \acrshort{rl} agent is shown by a green dashed line while persistent \acrshort{rl} agent is shown by green solid line. Baseline \acrshort{rl} is drawn with yellow line used for bench marking. Similar to the cart pole environment, both agents supported by the trainer, regardless of whether or not they used \acrshort{ppr}, obtain better results than baseline \acrshort{rl}. Then, the persistent agent achieves convergence results slightly earlier than its non-persistent counterpart. The trainer's accuracy and frequency feedback are used the same as in the cart pole environment, so the results are reflected for use in the domestic robot environment as well and, not just the ideal hypothetical environment like cart pole in AI gym.

Table \ref{table:wbfeedback-iteration} shows the average number and percentage of interactions that occurred for each agent. We can see that the number of interactions is similar in the experiment.

\begin{table}[ht]
    \begin{center}
        \begin{tabular}{lll}
        \toprule
        \textbf{Agent} & \multicolumn{2}{c}{\textbf{Interaction}} \\
        \cmidrule(r){2-3}
        & \textbf{Non-persistent} & \textbf{Persistent} \\
        \midrule
        Realistic Advisor   &   9077(47.64\%)  &   8241(47.18\%) \\
        \bottomrule
        \end{tabular}
    \end{center}
    \caption{The average number of interactions in experiment and the percent compare with total steps taken for realistic agent in Webots environment}
    \label{table:wbfeedback-iteration}
\end{table}

\begin{figure}[ht]
    \centering
    \begin{subfigure}[b]{0.9\linewidth}
        \includegraphics[width=\textwidth]{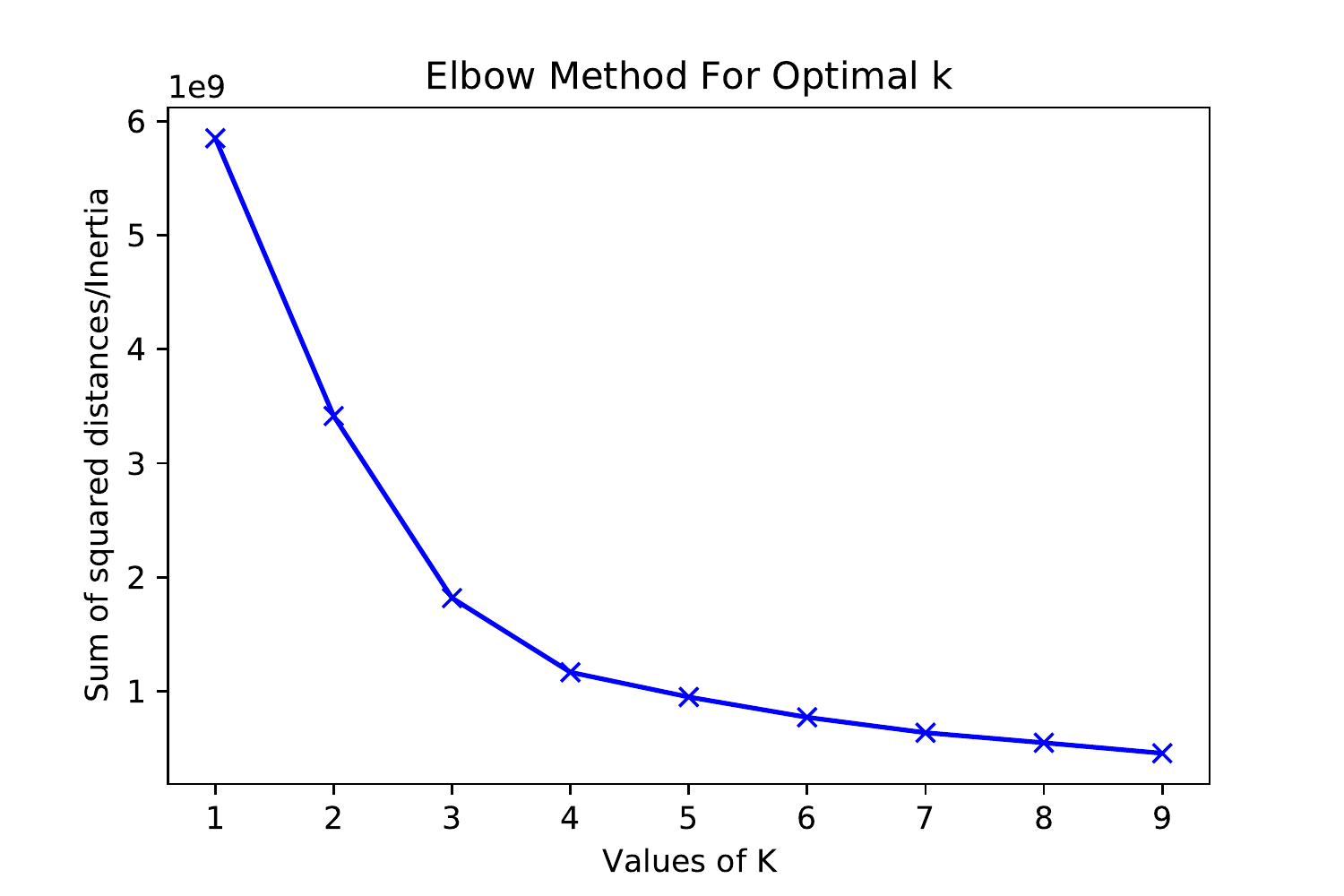}
        \caption{Elbow method}
        \label{fig:54a}
    \end{subfigure}
    \begin{subfigure}[b]{0.9\linewidth}
        \includegraphics[width=\textwidth]{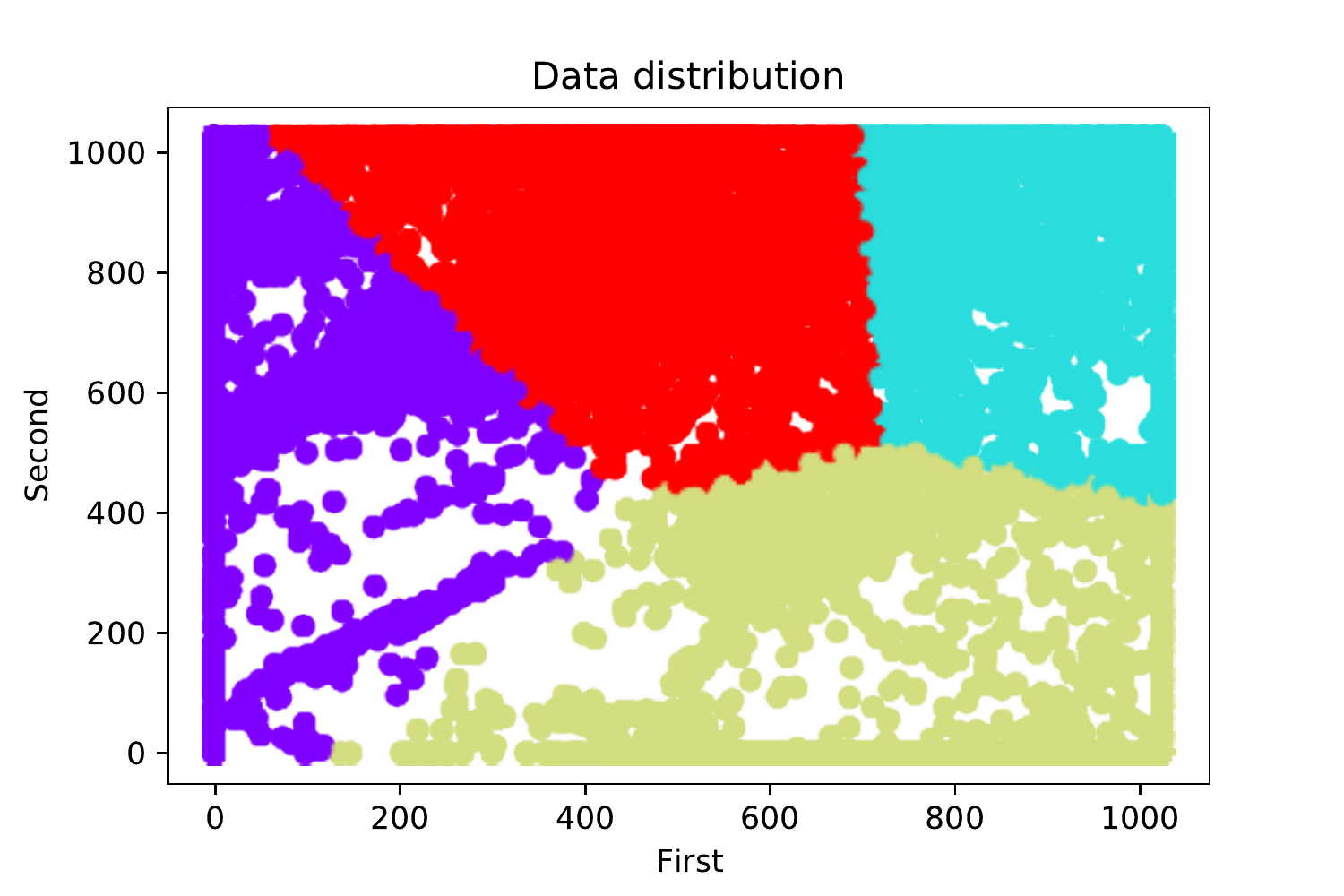}
        \caption{Distribution}
        \label{fig:54b}
    \end{subfigure}
    \caption{Total of squared distance for value k from 1-9 and distribution for 50000 states with value k = 4 at two value attributes of distance sensor.}
    \label{fig:54}
\end{figure}

Figure \ref{fig:54a} shows a graph using the elbow method to specify the number of clusters as a parameter of $k$-means, using the data of 50000 states in the actual running environment. The elbow method shows the best value for using k at the value 4.
Figure \ref{fig:54b} displays the data distribution in two axes of distance sensor value at the value k = 4.  

%% file: components/06_conclusion.tex
\section{Conclusion \& future work}
In this work, we proposed \acrshort{bpa}, a broad-persistent advising approach to implement the use of \acrshort{ppr} and generalised advice in continuous-state environments. Moreover, we also performed a comparison between autonomous \acrshort{drl}, \acrshort{dirl} without \acrshort{bpa} approach and \acrshort{dirl} with \acrshort{bpa} approach. Two environments were tested to investigate the impact that \acrshort{bpa} approach achieve the measured performance.

Overall, results obtained show that the \acrshort{bpa} approach with $k$-means as a generalise model and \acrshort{ppr} as a model of persistence performed slightly faster when the advice is withheld. The more accuracy for the advice and the longer time to retain it increase learning speed significantly.  Our research shows that the first step for applying PPR in a continuous state-space environment is feasible and effective. In addition, $k$-means used as broad advice inherits advantages based on its characteristic: runs fast, and scales very well over a large state space. Therefore, it is very suitable for the model to run in a real-world environment.

As future work, we are planning to further investigate the number of clusters of the $k$-means model which is obtained using the elbow method. Currently, it is quite small and, therefore, BPA is implemented to remember advice within a few steps. If agents try to keep in a long time with a small number of clusters will effectively affect the accuracy of the model suggested by the \acrshort{ppr}. The agent may find difficulties to achieve convergence results with many incorrect suggested actions. A more in-depth survey of the generalisation model is needed to get the best results for using \acrshort{ppr}. The accuracy of the generalisation model greatly affects the speed and convergence of the \acrshort{irl} model. Additionally, we suggest reducing the number of interactions with the trainer by reusing the action in the persistent model more often. When the agent reaches a new state that is already in memory, the agent reuses the recommended action immediately without interacting with the trainer. However, this should only be done when we have a good enough generalisation model. In addition, we plan to transfer and test the proposed approach on a real-world scenario with human and robot interaction.